\documentclass[10pt,twocolumn,letterpaper]{article}

\usepackage{cvpr}              

\usepackage{graphicx}
\usepackage{amsmath}
\usepackage{amssymb}
\usepackage{booktabs}
\usepackage{multirow}
\usepackage{bbding}
\usepackage{textcomp}
\usepackage{tablefootnote}
\usepackage[accsupp]{axessibility}

\usepackage{hyperref}
\hypersetup{pagebackref,breaklinks,colorlinks}

\usepackage[capitalize]{cleveref}
\crefname{section}{Sec.}{Secs.}
\Crefname{section}{Section}{Sections}
\Crefname{table}{Table}{Tables}
\crefname{table}{Tab.}{Tabs.}

\def\cvprPaperID{3113} 
\def\confName{CVPR}
\def\confYear{2023}

\begin{document}

\title{Quality-aware Pre-trained Models for Blind Image Quality Assessment}

\author{Kai Zhao$^{\dag}$, ~Kun Yuan$^{\dag}$, ~Ming Sun, ~Mading Li and Xing Wen \\
Kuaishou Technology \\
{\tt\small \{zhaokai05,yuankun03,sunming03,limading,wenxing\}@kuaishou.com}
}
\maketitle
\def\thefootnote{\dag}\footnotetext{Equal contribution.}

\begin{abstract}

Blind image quality assessment (BIQA) aims to automatically evaluate the perceived quality of a single image, whose performance has been improved by deep learning-based methods in recent years. However, the paucity of labeled data somewhat restrains deep learning-based BIQA methods from unleashing their full potential. In this paper, we propose to solve the problem by a pretext task customized for BIQA in a self-supervised learning manner, which enables learning representations from orders of magnitude more data. To constrain the learning process, we propose a quality-aware contrastive loss based on a simple assumption: the quality of patches from a distorted image should be similar, but vary from patches from the same image with different degradations and patches from different images. Further, we improve the existing degradation process and form a degradation space with the size of roughly $2\times10^7$. After pre-trained on ImageNet using our method, models are more sensitive to image quality and perform significantly better on downstream BIQA tasks. Experimental results show that our method obtains remarkable improvements on popular BIQA datasets. 

\end{abstract}


\section{Introduction}
\label{sec:intro}


With the arrival of the mobile internet era, billions of images are generated, uploaded and shared on various social media platforms, including Twitter, TikTok, \textit{etc} \cite{DBLP:journals/corr/abs-2011-00362}. As an essential indicator, image quality can help these service providers filter and deliver high-quality images to users, thereby improving Quality of Experience. Therefore, huge efforts \cite{DBLP:journals/tnn/GuTQL18, DBLP:journals/tip/WuMLDSL20, DBLP:journals/tip/RehmanW12, DBLP:journals/tip/GolestanehK16, DBLP:journals/tip/CiancioCSSSO11, DBLP:journals/tip/GhadiyaramB16} have been devoted to establishing an image quality assessment (IQA) method consistent with human viewers. In real-world scenarios, there usually exists no access to the reference images and the quality of reference images is suspicious. Thus, blind IQA (BIQA) methods are more attractive and applicable, despite full-reference IQA has achieved prospective results \cite{DBLP:conf/cvpr/Kim017}.

Recently, deep learning-based BIQA methods have made tremendous improvements on in-the-wild IQA benchmarks \cite{ DBLP:journals/tip/HosuLSS20, DBLP:conf/cvpr/FangZZMW20, DBLP:conf/cvpr/YingNGMGB20}. However, this problem is far from resolved and is hindered by the paucity of labeled data \cite{DBLP:journals/spm/KimZGLZB17}. The largest (by far) available BIQA dataset, FLIVE \cite{DBLP:conf/cvpr/YingNGMGB20}, contains 
nearly 40,000 real-world distorted images. By comparison, the very popular \textit{entry-level} image recognition dataset, CIFAR-100 \cite{krizhevsky2009learning}, contains 60,000 labeled images. Accordingly, existing BIQA datasets are too small to train deep learning-based models effectively.

\begin{figure}[t]
  \centering
    \includegraphics[width=0.85\linewidth]{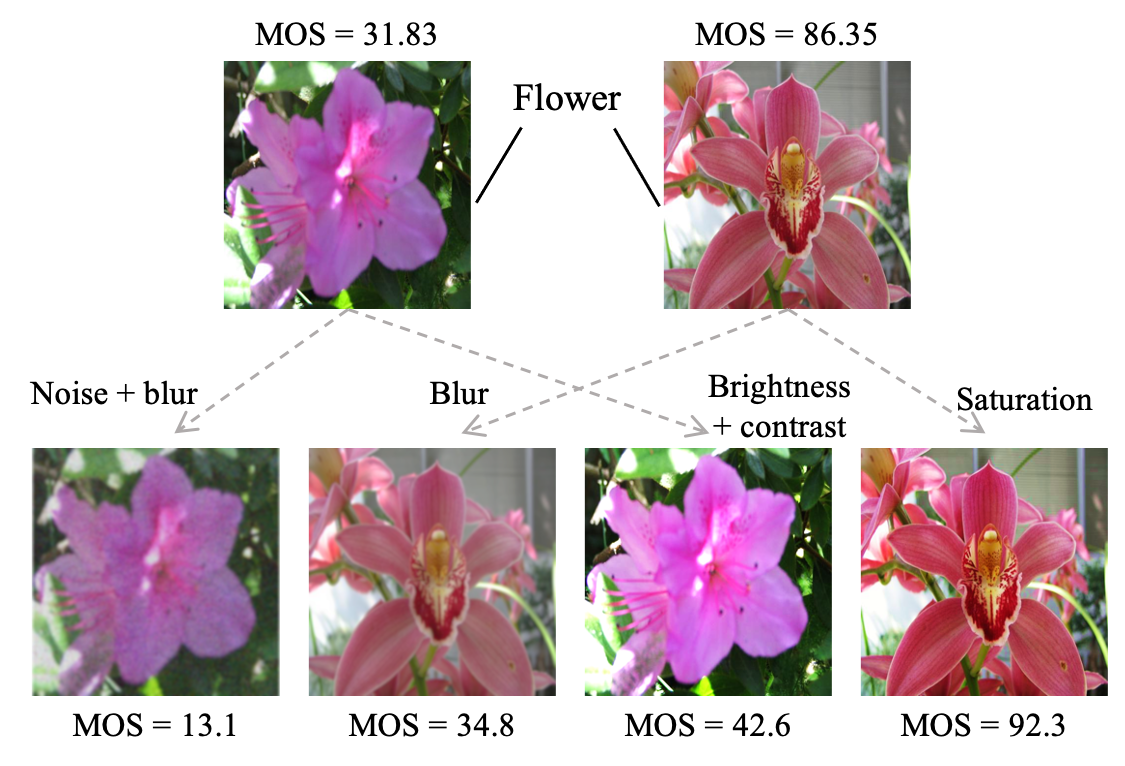}
  \caption{The two images in the first row are sampled from BIQA dataset CLIVE \cite{DBLP:journals/tip/GhadiyaramB16}. Although they have the same semantic meaning, their perceptual qualities are quite different: their mean opinion scores (MOS) are 31.83 and 86.35. The second row shows modified versions of the first two images and their MOSs (rated by 7 people). After different operations, the quality dramatically changed, while semantic meaning remains unchanged.}
  \label{fig:teaser}
  \vspace{-0.3cm}
\end{figure}

Researchers present several methods to tackle this challenge. A straight-forward way is to sample local patches and assign the label of the whole image (\ie, mean opinion score, MOS) to the patches \cite{DBLP:conf/cvpr/KangYLD14, DBLP:journals/tip/BosseMMWS18, DBLP:journals/tmm/LiJLJ19, DBLP:journals/tcsv/ZhangMYDW20, DBLP:conf/cvpr/SuYZZGSZ20, DBLP:journals/tip/ZhangMZY21, DBLP:conf/iccv/KeWWMY21}. However, the perceived scores of local image patches tend to differ from the score of the entire image \cite{DBLP:journals/tcsv/ZhangMYDW20, DBLP:conf/cvpr/YingNGMGB20}.
Another common strategy is to leverage domain knowledge from large-scale datasets (\eg, ImageNet \cite{DBLP:conf/cvpr/DengDSLL009}) for other computer vision tasks  \cite{DBLP:conf/iccv/KeWWMY21, DBLP:journals/tip/ChenLWDS22}. Nevertheless, these pre-trained models can be sub-optimal for BIQA tasks: images with the same content share the same semantic label, whereas, their quality may be different (\cref{fig:teaser}). 
Some researchers propose to train a model on synthetic images with artificial degradation and then regress the model onto small-scale target BIQA datasets\cite{DBLP:conf/icip/MaLFS19, DBLP:journals/tcsv/ZhangMYDW20, DBLP:journals/tip/WuMLDSL20}.
However, images generated by rather simple degradation process with limited distortion types/levels are far from authentic. Further, synthetic images are regularly distorted from limited pristine images of high quality, so the image content itself only has a marginal effect on the image quality. Yet in real-world scenarios, the image quality is closely related to its content, due to viewers' preferences for divergent contents \cite{DBLP:journals/tmm/LiJLJ19,  DBLP:journals/corr/abs-2105-14550}.



Self-supervised learning (SSL) or unsupervised learning is another potential choice to overcome the problem of lacking adequate training data, for its ability to utilize an amount of unlabeled data. Such technique is proven to be effective in many common computer vision tasks \cite{DBLP:journals/entropy/Albelwi22}. However, different from models for these tasks which mainly focus on high-level information, representations learned for BIQA should be sensitive to all kinds of low-level distortions and high-level contents, as well as interactions between them. There are few research focused on such area in the literature, let alone a deep SSL designed for BIQA with state-of-the-art performance \cite{DBLP:journals/tip/MadhusudanaBWAB22}.

In this work, we propose a novel SSL mechanism that distinguishes between samples with different perceptual qualities, generating \textit{Quality-aware Pre-Trained (QPT)} models for downstream BIQA tasks. Specifically, we suppose the quality of patches from a distorted image should be similar, but vary from patches from different images (\ie, \textit{content-based negative}) and the same image with different degradations (\ie, \textit{degradation-based negative}). Moreover, inspired by recent progress in image restoration, we introduce shuffle order \cite{DBLP:conf/iccv/0008LGT21}, high-order \cite{DBLP:conf/iccvw/WangXDS21} and a skip operation to the image degradation process, to simulate real-world distortions. In this way, models pre-trained on ImageNet are expected to extract quality-aware features, and boost downstream BIQA performances. We summarize the contributions of this work as follows:
\begin{itemize}
    \item We design a more complex degradation process suitable for BIQA. It not only considers a mixture of multiple degradation types, but also incorporates with shuffle order, high-order and a skip operation, resulting in a much larger degradation space. For a dataset with a size of $10^7$, our method can generate more than $2\times10^{14}$ possible pairs for contrastive learning.
    \item To fully exploit the abundant information hidden beneath such an amount of data,
    we propose a novel SSL framework to generate QPT models for BIQA based on MoCoV2 \cite{DBLP:journals/corr/abs-2003-04297}. By carefully designing positive/negative samples and customizing quality-aware contrastive loss, our approach enables models to learn quality-aware information rather than regular semantic-aware representation from massive unlabeled images.
    \item Extensive experiments on five BIQA benchmark datasets (sharing the same pre-trained weight of QPT) demonstrate that our proposed method significantly outperforms other counterparts, which indicates the effectiveness and generalization ability of QPT. 
    It is also worth noting that the proposed method can be easily integrated with current SOTA methods by replacing their pre-trained weights.
    
    
\end{itemize}

\section{Related Work}
In this section, we provide a brief review of major works on BIQA and recent advances in contrastive learning and image degradation modeling. 

\subsection{Blind Image Quality Assessment}
Before the rise of deep learning, natural scene statistics (NSS) theory dominates the realm of BIQA, which assumes that pristine natural images obey certain statistics distribution and various distortions will break such statistical regularity  \cite{simoncelli2001natural}. Based on this theory, various hand-crafted features are proposed in different domains, including spatial \cite{DBLP:journals/tip/MittalMB12, DBLP:journals/spl/MittalSB13}, frequency \cite{DBLP:journals/tip/SaadBC12, DBLP:journals/tip/MoorthyB11} and gradient \cite{DBLP:journals/tip/ZhangZB15}. Meanwhile, some learning-based methods \cite{ DBLP:conf/cvpr/YeKKD12, DBLP:journals/tbc/MinZGLY18} have also been preliminarily explored. Typically, they estimate the subjective quality using support vector regression and features learned by visual codebooks \cite{DBLP:journals/tip/YeD12}.

In recent years, a variety of deep learning-based methods have been studied for BIQA and significantly boost the in-the-wild performance\cite{DBLP:conf/cvpr/HeZRS16, DBLP:conf/nips/RenHGS15, DBLP:conf/nips/VaswaniSPUJGKP17, DBLP:journals/tmm/LiJLJ19, DBLP:conf/wacv/GolestanehDK22}. As a pioneer work, a shallow network \cite{DBLP:conf/cvpr/KangYLD14} consisting of only three layers is built to solve BIQA in an end-to-end manner. After that, later works naturally expand the BIQA model by deepening the network depth \cite{DBLP:journals/tip/BosseMMWS18, DBLP:journals/tip/MaLZDWZ18} or utilizing more effective building blocks \cite{DBLP:conf/cvpr/SuYZZGSZ20, DBLP:conf/cvpr/ZhuLWDS20}. Recently, transformer-based BIQA methods \cite{DBLP:conf/icip/YouK21, DBLP:conf/iccvw/ZhuHCXLC21, DBLP:conf/iccv/KeWWMY21, DBLP:conf/cvpr/YangWSLGCWY22} 
are sprouting up, based on the assumption that transformer can compensate the missing ability of CNNs to capture non-local information \cite{DBLP:conf/iclr/DosovitskiyB0WZ21}.

Except for elaborate model design, a few research are devoted to solving the chief obstacle in BIQA, the paucity of labeled data. Some of them try to make full use of existing supervisory signals, such as rank learning \cite{DBLP:conf/iccv/LiuWB17, DBLP:conf/mm/LiJJ20},  multi-task learning \cite{DBLP:conf/cvpr/FangZZMW20} and mixed-dataset training \cite{DBLP:journals/corr/abs-2105-14550, DBLP:journals/tip/ZhangMZY21}. Meanwhile, others turn to large-scale pre-training \cite{DBLP:journals/tip/MaLZDWZ18, DBLP:conf/icip/ZengZB18, DBLP:conf/iccv/KeWWMY21}, which generates a large number of distorted images labeled by specifications of degradation process \cite{DBLP:journals/tcsv/ZhangMYDW20} or quality scores estimated from FR models \cite{DBLP:conf/icip/MaLFS19}. 

In comparison, our proposed method is based on contrastive learning without needing any explicit supervisory signal, thereby fully utilizing plenty of real-world images.

\subsection{Self-supervised Learning}
SSL, as a form of unsupervised learning, is used to learn a good data representation for the downstream true purpose \cite{DBLP:journals/entropy/Albelwi22}. A very straightforward idea to conduct such learning is minimizing the difference between a model's output and a \textit{fixed} target, such as reconstructing the input pixels \cite{DBLP:journals/jmlr/VincentLLBM10} or predicting the pre-defined categories \cite{DBLP:conf/iccv/DoerschGE15, DBLP:conf/eccv/ZhangIE16}. Motivated by the successes of masked language modeling in natural language processing, masked image modeling \cite{DBLP:conf/cvpr/HeCXLDG22, DBLP:conf/cvpr/Xie00LBYD022} is becoming a hot trend in computer vision.

Unlike the above methods, the goal of contrastive learning is to learn such an embedding space, where similar samples are assembled while dissimilar pairs are repelled \cite{DBLP:journals/corr/abs-2011-00362}. Specifically, contrastive learning generally involves two aspects: pretext tasks and training objectives. Owing to the flexibility of contrastive learning, a wide range of pretext tasks have been proposed, such as colorization \cite{DBLP:conf/eccv/ZhangIE16, DBLP:conf/cvpr/ZhangIE17}, multi-view coding \cite{DBLP:conf/eccv/TianKI20}, and instance discrimination \cite{DBLP:conf/cvpr/WuXYL18, DBLP:conf/cvpr/He0WXG20}. When it comes to training objectives, the major trend is the move from only one positive and one negative sample \cite{DBLP:conf/cvpr/ChopraHL05, DBLP:conf/cvpr/SchroffKP15} to multiple positive and negative pairs \cite{DBLP:journals/jmlr/GutmannH10, DBLP:journals/corr/abs-1807-03748, DBLP:conf/icml/FrosstPH19}. Especially, InfoNCE \cite{DBLP:journals/corr/abs-1807-03748} has become the most popular choice \cite{DBLP:conf/cvpr/He0WXG20, DBLP:conf/nips/CaronMMGBJ20, DBLP:conf/iccv/ChenXH21}. 

Notably, all of the above works can be classified as \textit{semantic-aware} pre-training, because they encourage views (augmentations) of the same image to have a similar representation while ignoring the changes of perceived image quality. In this work, we redesign a quality-aware pretext task for BIQA, which will be discussed in \cref{sec:pretext}.

\subsection{Image Degradation Modelling}
Most of the previous works focus on several classical distortion types \cite{DBLP:journals/corr/abs-1208-3718, DBLP:conf/nips/WangS03a, DBLP:conf/icip/SheikhWB02, DBLP:journals/tip/SheikhBC05} to extract distortion-specific features. As the importance of BIQA is rising, more distortion types have been developed and applied to synthesize distorted images \cite{DBLP:journals/tip/MaDWWYLZ17}. DipIQ \cite{DBLP:journals/tip/MaLLWT17} summarizes common distortion operations, such as noise, blur and compression, and further subdivides each with five levels. DB-CNN \cite{DBLP:journals/tcsv/ZhangMYDW20} introduces additional five distortion types. Recently, many works \cite{DBLP:journals/spm/KimZGLZB17, DBLP:conf/cvpr/ZhuLWDS20, DBLP:conf/wacv/GolestanehDK22} point out that the above methods are too simple to simulate real-world images with complex and mixed distortions, due to the synthesized image only with a specific distortion type and level. 
In this work, we combine multiple degradation tricks \cite{DBLP:conf/iccv/0008LGT21,  DBLP:conf/iccvw/WangXDS21} proposed in the field of image restoration to form a much larger degradation space and generate distorted images that are more realistic.



\section{Method}

To excavate the potential of the ``pre-training and fine-tuning'' paradigm on BIQA for better performance, we present QPT models by pre-training on ImageNet. 
In \cref{sec:moco}, we will briefly review some related SSL frameworks. 
In \cref{sec:op}, we will detail the degradation process for generating distorted images that contain quality-related information. 
In \cref{sec:pretext}, we will give the definition of quality-aware pretext task and optimization objective.

\begin{figure*}[t]
  \centering
    \includegraphics[width=1.0\linewidth]{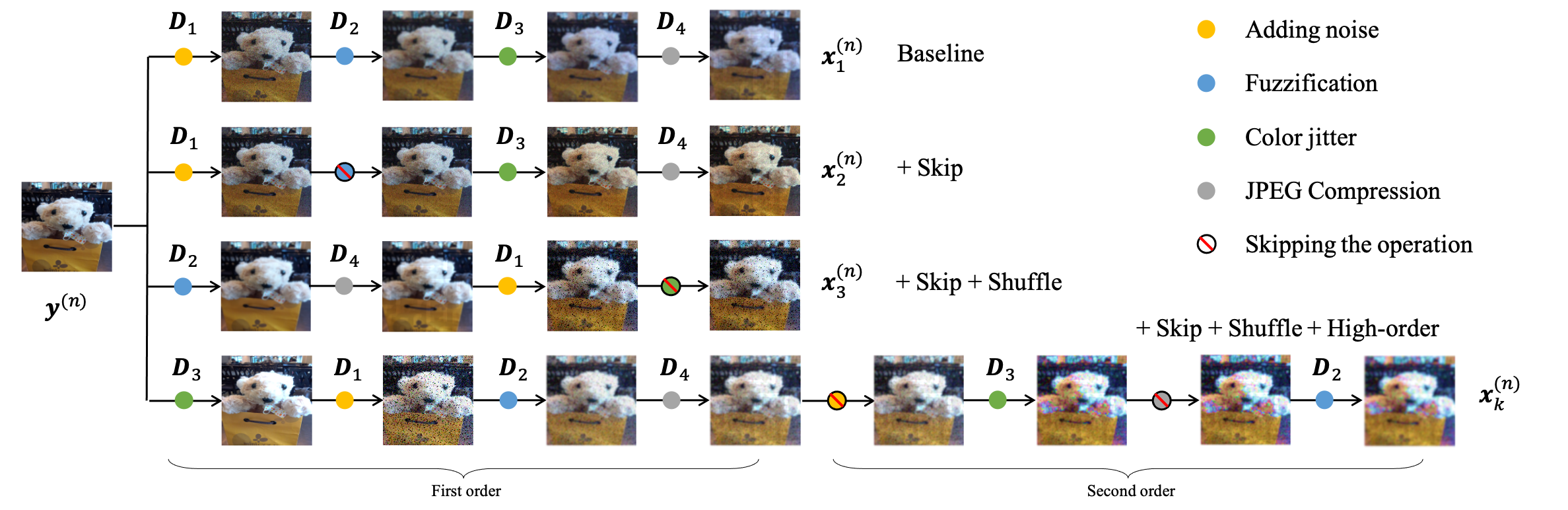}
  \caption{Illustration of generating distorted images using different compositions of degradation. Compared with the process of fixed sequence, the introduced \textit{skip, shuffle and high-order} largely increase the degradation space, covering diverse and realistic distortions.}
  \label{fig:aug}
  \vspace{-0.3cm}
\end{figure*}

\subsection{Reviewing the Framework of SSL}\label{sec:moco}

SSL enables learning representations from orders of magnitude more data, which addresses the obstacle for BIQA in establishing large-scale annotated datasets. SSL methods can be roughly divided into generative-based \cite{DBLP:conf/icml/VincentLBM08,DBLP:conf/cvpr/HeCXLDG22,DBLP:conf/cvpr/Xie00LBYD022} and contrastive-based \cite{DBLP:conf/cvpr/He0WXG20,DBLP:conf/icml/ChenK0H20,DBLP:conf/nips/GrillSATRBDPGAP20}. Among them, MAE \cite{DBLP:conf/cvpr/HeCXLDG22}, as a generative-based method, develops an asymmetric encoder-decoder architecture for image reconstruction. MoCo \cite{DBLP:conf/cvpr/He0WXG20}, as a contrastive-based method, builds a dynamic dictionary with a queue and a moving-averaged encoder for contrastive learning. Relatively, contrastive learning is more suitable for BIQA, for which can easily measure quality ranking between samples. Due to the fact that the quality of an image is related to various complicated factors \cite{DBLP:journals/tip/TuWBAB21,DBLP:conf/cvpr/WangKTYBAMY21}, a large number of images are needed to ensure the diversity of negative samples \cite{DBLP:conf/icml/ChenK0H20}. Empirically, we choose MoCo, since it provides a memory-efficient framework to establish a large and consistent set.

In brief, MoCo builds a dictionary using a momentum queue to enlarge available size for comparison, where samples are progressively replaced with incoming mini-batch ones. And keys in the dictionary are updated through a momentum-based moving average of the encoder network, maintaining consistency with query samples. MoCo adopts the instance-level discrimination task \cite{DBLP:conf/cvpr/WuXYL18} as its pretext task (\ie, a query matches a key if they are encoded views of the same image). Then, a contrastive loss, named InfoNCE, is adopted to distinguish the positive key from negative ones in a softmax manner, which can be noted as:
\begin{equation}
    \mathcal{L}_{\text{InfoNCE}} = -\log \frac{\exp(q\cdot k_{+}/\tau)}{\sum_{i=0}^N \exp(q\cdot k_i/\tau)},
    \label{infonce}
\end{equation}
where $q$ is an encoded query, $k_i$ is the $i$-th encoded key of a dictionary consisting of $N$ samples, and $k_{+}$ is a single key that matches $q$. And $\tau$ is a temperature hyper-parameter.

\subsection{Degradation Space}\label{sec:op}




To obtain utilizable quality-related information in self-supervised scenarios, a simple way is to generate image pairs using controllable degradations manually. 
Before designing specific degradation types, several observations need to be mentioned:
\textbf{First}, there exist many factors affecting the perceptual quality of an image (\eg, content, distortions, compression). For example, an image with unmeaningful content is often classified as a low-quality one \cite{DBLP:conf/mm/LiJJ19}. Also, distortions (\eg, blurriness \cite{DBLP:journals/sivp/CohenY10}, noise \cite{DBLP:journals/tip/ShenLE11} and sharpness \cite{DBLP:journals/tip/YueHGZZ19}) introduced during the image production phase and compression artifacts introduced by transcoding or transmission, will degrade the overall quality.
\textbf{Second}, the aforementioned factors, as a complicated composition, are always involved in practical scenarios \cite{DBLP:journals/tip/ShenLE11,DBLP:conf/cvpr/AhnCY21}. An image may go through various processes of shooting, editing, compression, and transmission, resulting in more complicated distortions.
These scenarios increase the difficulty of BIQA tasks. Thus a good pre-trained model for IQA should take these into account, covering diverse distortions using suitable degradations as large and realistic as possible.

Based on these observations, we design the degradation space from the perspective of individual operations and their compositions. \textbf{For the first observation}, we introduce a variety of degradation types to simulate real distortions, which can be divided into three categories: (1) {\textit{geometric deformation}}, which simulates distortions introduced during editing process or adaptation on various display equipment, including 4 operations of \underline{scale jitter} \cite{DBLP:conf/iccv/KeWWMY21}, \underline{horizontal flip}, \underline{down-sampling} and \underline{up-sampling} \cite{DBLP:conf/icip/OuWZ19}; (2) \textit{color change} \cite{DBLP:journals/spic/TemelA16}, which can be caused by lightness, chroma, and hue differences in the shooting or codec process, including 2 operations of \underline{color jitter} and \underline{grayscale}; (3) \textit{texture adjustment}, which can be captured from environmental interference or transmission, including 3 operations of \underline{adding noise}, \underline{fuzzification} \cite{DBLP:conf/mm/GaoMZL0Z22} and \underline{JPEG compression} \cite{DBLP:journals/bspc/ChowP16}. Given an image $\mathbf{y}$, the process can be formulated as:
\begin{equation}
    \mathbf{x} = \mathcal{D}(\mathbf{y};w,p),
\end{equation}
where $\mathcal{D}$ denotes the function of degradation. $w$ represents hyper-parameters of the degradation (\eg, scale size, location, brightness intensity, \textit{etc.}). And $p\in \{0,1\}$ indicates whether this operation is performed. \textbf{For the second observation},  we propose a sequence, consisting of \textit{randomly chosen} degradations, for more complicated types (\eg, a list of \{scale jitter, color jitter, adding noise\}), where each operation can be \textit{skipped}, and the order can be \textit{shuffled}:
\begin{equation}\label{equ:degradation}
    \mathbf{x} = \mathcal{D}^o(\mathbf{y}) = \mathcal{D}_o(\cdots \mathcal{D}_1(\mathbf{y};w_1,p_1)\cdots ;w_o,p_o),
\end{equation}
where $o$ is the number of selected degradations. Furthermore, to simulate multiple transfers, we adopt a \textit{high-order} degradation type, involving multiple repeated processes of $\mathcal{D}^o(\mathbf{x})$, where each process performs with the same procedure but different hyper-parameters. Empirically, we employ a degenerate approach with at most two orders, as it could resolve most real cases while keeping simplicity. The illustration of the overall process is shown in \cref{fig:aug}.

The degradation space formed by our operations is much larger than those in existing super-resolution (SR) methods \cite{DBLP:conf/iccvw/WangXDS21,DBLP:conf/iccv/0008LGT21}, which reconstruct high-resolution images from low-resolution ones. Since BIQA does not require pixel-level aligned pairs like SR, more operations like geometric deformation/color change can be used. Moreover, SR requires that the source images should be high-quality and the distributions of target and source images should be as consistent as possible. These factors limit the number of available compositions. Without these restrictions, in our method, the size of degradation space can be expanded substantially. Theoretically, for a space containing 9 degradation types, the number of discrete compositions is  $2\times\sum_{i=1}^9\mathbf{C}_9^i\times\mathbf{A}_i^i \thickapprox 2\times 10^{7}$.

\begin{figure*}[t]
  \centering
    \includegraphics[width=1.0\linewidth]{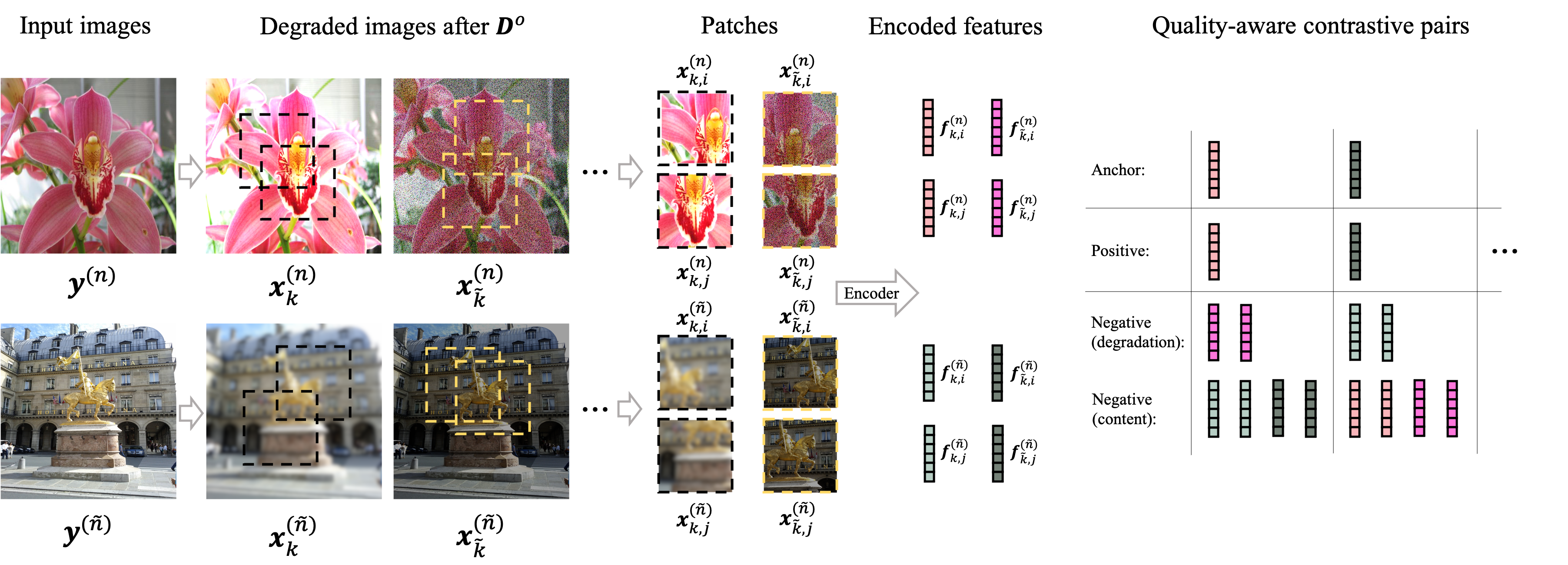}
    \vspace{-0.5cm}
    \caption{The framework of \textit{Quality-aware Pretext Task (QPT)}. First, images are augmented by diverse degradations. Second, patches will be extracted from various views to form patch pairs. Pairs generated from the same view but different locations are treated as positive pairs. And other pairs are noted as negative pairs, including degradation-based and content-based. Then, the patches are transformed into encoded features by a network. Last, these feature pairs are constrained using a \textit{Quality-aware Contrastive Loss (QC-Loss)}.}
    \vspace{-0.3cm}
  \label{fig:framework}
\end{figure*}

\subsection{Quality-aware Pretext Task}\label{sec:pretext}




Different from the classical \textit{semantic-aware} pretext task (\ie, treating each image instance as a distinct class) \cite{DBLP:conf/cvpr/WuXYL18,DBLP:conf/nips/BachmanHB19}, we propose a new \textit{quality-aware} pretext task that distinguishes between samples with different perceptual qualities.


Suppose we have $N$ images $\mathbf{y}^{(1)}, \cdots, \mathbf{y}^{(N)}$. As shown in \cref{fig:framework}, each image $\mathbf{y}^{(n)}$ will be differently augmented by $K$ degradation compositions, obtaining different ``views'' $\mathbf{x}^{(n)}_1, \cdots, \mathbf{x}^{(n)}_K$ that own consistent content but with different perceptual qualities. 
Then we extract a ``query'' (or ``anchor'') and a ``key'' patch that originates from the same view $\mathbf{x}^{(n)}_k$ but with different locations, denoted $\mathbf{x}^{(n)}_{k,i}$ and $\mathbf{x}^{(n)}_{k,j}$ respectively. 
The query and key are considered as a positive pair due to their consistent content and degradation type.
And otherwise pairs are treated as negative ones, which can be divided into two parts.
First, for the same image, patches extracted from different ``views'' are denoted as \textbf{degradation-based} negative pairs. Their content is the same, but the way it degenerates is different, resulting in inconsistent qualities. To ensure the quality of patches generated from a single image as consistent as possible, we set the lower bound for the area ratio during random crop to be 0.5. So patches generated from the same image share similar content rather than contrast regions (\eg, smooth and textured ones).
Besides, for different images, regardless of whether their degradation types are the same, they are considered \textbf{content-based} negative pairs due to different contents. 
Notably, there may exist some noise (\ie, different contents, similar quality) under this partition, but it is negligible due to the very small proportions. To count the number of negative pairs with similar quality, we perform JND tests proposed in LPIPS \cite{DBLP:conf/cvpr/ZhangIESW18} on 500 generated patches, indicating around 10\%. Correctly assigned pairs still dominate. The difference between quality-level and instance-level discrimination tasks is given in \cref{tab:pretext}.

\begin{table}[t]
    \centering
    \footnotesize
    \caption{Comparison of two types of pretext tasks, where $\tilde{k}$ represents a degradation/augmentation type different from $k$, $\tilde{n}$ is an image different from $n$, and $\ast$ denotes an arbitrary value.}
    \begin{tabular}{c|c|c}
    \toprule
        Pretext task & Positive Pair & Negative Pair \\
    \midrule
        \textit{semantic-aware} & $(\mathbf{x}^{(n)}_{k,i},\mathbf{x}^{(n)}_{\tilde{k},j})$ & $(\mathbf{x}^{(n)}_{\ast,\ast},\mathbf{x}^{(\tilde{n})}_{\ast,\ast})$ \\
    \midrule
        \textit{quality-aware} & $(\mathbf{x}^{(n)}_{k,i},\mathbf{x}^{(n)}_{k,j})$ & $(\mathbf{x}^{(n)}_{\ast,\ast},\mathbf{x}^{(\tilde{n})}_{\ast,\ast})$,  $(\mathbf{x}^{(n)}_{k,\ast},\mathbf{x}^{(n)}_{\tilde{k},\ast})$ \\
    \bottomrule
    \end{tabular}
    \label{tab:pretext}
    \vspace{-0.4cm}
\end{table}

Let $\mathcal{F}$ denote the transformation of the encoder network. Given an input patch $\mathbf{x}^{(n)}_{k,i}$, the encoded feature can be noted as $\mathbf{f}^{(n)}_{k,i} = \mathcal{F}(\mathbf{x}^{(n)}_{k,i})/||\mathcal{F}(\mathbf{x}^{(n)}_{k,i})||$, after a $\mathcal{L}_2$ normalization for computing the dot-product similarity. Then our method can be optimized using a \textit{Quality-aware Contrastive Loss (QC-Loss)}, which is formulated as follows:
\begin{equation}
\small
\begin{aligned}
    \mathcal{L}_{\text{QC}} =  & - \beta \cdot \log \sum_{k=1}^K \frac{ \exp(\mathbf{f}^{(n)}_{k,i}\cdot \mathbf{f}^{(n)}_{k,j}/\tau)}
    {
    \sum_{\tilde{k}\neq k}^K \exp(\mathbf{f}^{(n)}_{k,i}\cdot \mathbf{f}^{(n)}_{\tilde{k},j}/\tau)
    } \\
    & - \log \sum_{k=1}^K 
    \frac{\exp(\mathbf{f}^{(n)}_{k,i}\cdot \mathbf{f}^{(n)}_{k,j}/\tau)}
    {\sum_{\tilde{n}\neq n}^N \sum_{k^{\prime}=1}^K \exp(\mathbf{f}^{(n)}_{k,i}\cdot \mathbf{f}^{(\tilde{n})}_{k^{\prime},j}/\tau)},
    \label{qcLoss}
\end{aligned}
\end{equation}
where the first part denotes distinguishing degradation-based negative pairs, and the second part is for content-based negative pairs. $\beta$ is a coefficient for balancing.

To further construct more image pairs containing abundant content and texture information, we adopt the well-known ImageNet as the baseline dataset for pre-training. Compared with IQA datasets, whose number of available data is relatively small (as shown in \cref{tab:dataset}), ImageNet contains over 1 million of natural images from 1,000 diverse categories. Numerous samples and huge degradation space together contribute to generating $2\times 10^{14}$ possible pairs during the pre-training phase.

\section{Experiments}

\subsection{Datasets and Evaluation Criteria}

\paragraph{Datasets.} Our method is evaluated on five public BIQA datasets, including BID \cite{DBLP:journals/tip/CiancioCSSSO11}, CLIVE \cite{DBLP:journals/tip/GhadiyaramB16}, KonIQ10K \cite{DBLP:journals/tip/HosuLSS20}, SPAQ \cite{DBLP:conf/cvpr/FangZZMW20} and FLIVE \cite{DBLP:conf/cvpr/YingNGMGB20}. BID is a blur image dataset that contains 586 images with realistic blur distortions (\eg, out-of-focus, simple motion, complex motion blur). CLIVE consists of 1,162 images with diverse authentic distortions captured by mobile devices. KonIQ10K contains 10,073 images which are selected from YFCC-100M. And the selected images cover a wide and uniform range of distortions in
terms of quality indicators such as brightness, colorfulness, contrast, noise, sharpness, \textit{etc}. SPAQ consists of 11,125 images captured by different mobile devices, covering a large variety of scene categories. FLIVE is the largest in-the-wild IQA dataset by far, which contains 39,810 real-world distorted images with diverse contents, sizes and aspect ratios. Details are given in \cref{tab:dataset}.

\paragraph{Evaluation criteria.} Pearson’s Linear Correlation Coefficient (PLCC) and Spearman’s Rank-Order Correlation Coefficient (SRCC) are selected as criteria to measure the accuracy and monotonicity, respectively. They are in the range of [0, 1]. A larger PLCC means a more accurate numerical fit with MOS scores. A larger SRCC shows a more accurate ranking between samples. For all datasets, following \cite{DBLP:conf/iccv/KeWWMY21, DBLP:conf/wacv/GolestanehDK22}, we split
the dataset into a 80\% training set and a 20\% testing set randomly.

\begin{table}[t]
  \centering
  \footnotesize
  \caption{Details of selected BIQA datasets.}
  \begin{tabular}{c|ccc}
    \toprule
    Dataset & Size & Resolution & MOS Range\\
    \midrule
    BID~\cite{DBLP:journals/tip/CiancioCSSSO11} & 586 & 480P$\sim$2112P & 0$\sim$5 \\
    CLIVE~\cite{DBLP:journals/tip/GhadiyaramB16} & 1,162  & 500P$\sim$640P & 0$\sim$100 \\
    KonIQ10K~\cite{DBLP:journals/tip/HosuLSS20} & 10,073 & 768P & 0$\sim$100 \\
    SPAQ~\cite{DBLP:conf/cvpr/FangZZMW20} & 11,125 & 1080P$\sim$4368P & 0$\sim$100 \\
    FLIVE~\cite{DBLP:conf/cvpr/YingNGMGB20} & 39,810 & 160P$\sim$700P & 0$\sim$100 \\
    \bottomrule
  \end{tabular}
  \label{tab:dataset}
  \vspace{-0.3cm}
\end{table}

\begin{table*}[t]
  \centering
  \footnotesize
  \caption{Performance of existing SOTA methods and the proposed QPT models on five in-the-wild BIQA datasets. The ``$\star$" means missing corresponding results in the original paper. The best and second-best results are \textbf{bolded} and \underline{underlined}. With the inherent learned knowledge in QPT models, a widely-used ResNet50 can outperform existing SOTA methods.}
  \begin{tabular}{c|cc|cc|cc|cc|cc}
    \toprule
    \multirow{2}{*}{Method} & \multicolumn{2}{c|}{BID} & \multicolumn{2}{c|}{CLIVE} & \multicolumn{2}{c|}{KonIQ10K} & \multicolumn{2}{c|}{SPAQ} & \multicolumn{2}{c}{FLIVE} \\
    & SRCC & PLCC & SRCC & PLCC & SRCC & PLCC & SRCC & PLCC & SRCC & PLCC \\ 
    \midrule
    NIQE~\cite{DBLP:journals/spl/MittalSB13} & 0.4772 & 0.4713 & 0.4536 & 0.4676 & 0.5260 & 0.4745 & 0.6973 & 0.6850 & 0.1048 & 0.1409 \\
    ILNIQE~\cite{DBLP:journals/tip/ZhangZB15}  & 0.4946 & 0.4538 & 0.4531 & 0.5114 & 0.5029 & 0.4956 & 0.7194 & 0.654 & 0.2188 & 0.2547 \\
    BRISQUE~\cite{DBLP:journals/tip/MittalMB12} & 0.5736 & 0.5401 & 0.6005 & 0.6211 & 0.7150 & 0.7016 & 0.8021 & 0.8056 & 0.3201 & 0.3561 \\
    BMPRI~\cite{DBLP:journals/tbc/MinZGLY18} & 0.5154 & 0.4583 & 0.4868 & 0.5229 & 0.6577 & 0.6546 & 0.7501 & 0.7544 & 0.2737 & 0.3146 \\
    CNNIQA~\cite{DBLP:conf/cvpr/KangYLD14} & 0.6163 & 0.6144 & 0.6269 & 0.6008 & 0.6852 & 0.6837 & 0.7959 & 0.7988 & 0.3059 & 0.2850 \\
    WaDIQaM-NR~\cite{DBLP:journals/tip/BosseMMWS18} & 0.6526 & 0.6359 & 0.6916 & 0.7304 & 0.7294 & 0.7538 & 0.8397 & 0.8449 & 0.4346 & 0.4303 \\
    SFA~\cite{DBLP:journals/tmm/LiJLJ19} & 0.8202 & 0.8253 & 0.8037 & 0.8213 & 0.8882 & 0.8966 & 0.9057 & 0.9069 & 0.5415 & 0.6260 \\
    DB-CNN~\cite{DBLP:journals/tcsv/ZhangMYDW20} & 0.8450 & 0.8590 & 0.8443 & 0.8624 & 0.8780 & 0.8867 & 0.9099 & 0.9133 & 0.5537 & 0.6518 \\
    HyperIQA~\cite{DBLP:conf/cvpr/SuYZZGSZ20} & 0.8544 & 0.8585 & \underline{0.8546} & 0.8709 & 0.9075 & 0.9205 & 0.9155 & 0.9188 & 0.5354 & 0.6228 \\
    PaQ-2-PiQ$^{\star}$~\cite{DBLP:conf/cvpr/YingNGMGB20} & - & - & 0.840 & 0.850 & 0.870 & 0.880 & - & - & 0.571 & 0.623 \\
    CONRTIQUE$^{\star}$~\cite{DBLP:journals/tip/MadhusudanaBWAB22} & - & - & 0.854 & 0.890 & 0.896 & 0.901 & - & - & \underline{0.580} & 0.641 \\
    UNIQUE$^{\star}$~\cite{DBLP:journals/tip/ZhangMZY21} & \underline{0.858} & \underline{0.873} & \underline{0.854} & \underline{0.890} & 0.896 & 0.901 & - & - & - & - \\
    MUSIQ$^{\star}$~\cite{DBLP:conf/iccv/KeWWMY21} & - & - & - & - & \underline{0.916} & \underline{0.928} & \underline{0.917} & \underline{0.921} & {0.566} & \underline{0.661} \\
    TReS$^{\star}$~\cite{DBLP:conf/wacv/GolestanehDK22}  & - & - & 0.846 & 0.877 & 0.915 & \underline{0.928} & - & - & 0.554 & 0.625 \\
    \midrule
    ResNet50 & 0.8423 & 0.8521 & 0.8527 & 0.8807 & 0.9121 & 0.9270 & 0.9161 & 0.9207 & 0.5514 & 0.6354\\ 
    QPT-ResNet50 & \textbf{0.8875} & \textbf{0.9109} & \textbf{0.8947} & \textbf{0.9141} & \textbf{0.9271} & \textbf{0.9413} & \textbf{0.9250} & \textbf{0.9279} & \textbf{0.6104}\tablefootnote{Results in FLIVE follow an aligned train/test split with CONTRIQUE. When using random 80\%/20\% splits, SRCC is 0.5746, PLCC is 0.6748.} & \textbf{0.6770} \\
    \bottomrule
  \end{tabular}
  \label{tab:sota}
\end{table*}

\subsection{Implementation Details}
Our experiments are performed using PyTorch \cite{DBLP:conf/nips/PaszkeGMLBCKLGA19}, and are all conducted on 8 Nvidia V100 GPUs. Following the classical pre-training and fine-tuning paradigm, we run the whole experiments in two stages:

\paragraph{Pre-training.} We inherit most settings from MoCoV2 \cite{DBLP:journals/corr/abs-2003-04297}, while modify the pretext task and degradation process for the quality-aware purpose. Specifically, we use a ResNet50 \cite{DBLP:conf/cvpr/HeZRS16} with an extra two-layer MLP as the encoder and train it on ImageNet. We use SGD optimizer with weight decay of 0.0001, momentum of 0.9 and a mini-batch size of 256. The initial learning rate is 0.03 and multiplied by 0.1 at 120 and 160 epoch. The hyper-parameters, $\beta$ and $\tau$ in \cref{qcLoss}, are empirically set to 0.4 and 0.2 in our experiments. It takes about 75 hours training ResNet50, for 200 epochs. 

\paragraph{Fine-tuning.} Following the scheme from \cite{DBLP:journals/corr/abs-2105-14550}, we resize the smaller edge of images to 340 while maintaining aspect ratio, and randomly sample and horizontally flip patches with size 320x320 pixels. We use AdamW optimizer with weight decay of 0.01 and mini-batch size of 64. The Learning rate is initialized with 0.0002, and decayed by the cosine annealing strategy. For BID and CLIVE, experiments are trained for 100 epochs. And for other three dataset, the number is 200. By default, we select last epoch for weight initialization and evaluation. During test stage, we take five crops (\ie, four corners and a center patch) from an image, and average their corresponding prediction scores to get the final prediction score. To reduce the randomness in training/test set splitting, following \cite{DBLP:conf/wacv/GolestanehDK22, DBLP:conf/cvpr/SuYZZGSZ20, DBLP:conf/cvpr/ZhuLWDS20}, we repeat train/test procedures 10 times for each dataset, and report the median SRCC and PLCC results.

\begin{table*}[t]
    \centering
    \footnotesize
    \caption{Improving the performance of existing SOTA methods by replacing their pre-trained weights with QPT models directly. Experiments are conducted based on their open-source code, using the consistent dataset partitioning and training hyper-parameters.}
    \begin{tabular}{c|c|ll|ll|ll}
    \toprule
    \multirow{2}{*}{Methods} & \multirow{2}{*}{Pre-trained Type} & \multicolumn{2}{c|}{BID} & \multicolumn{2}{c|}{CLIVE} & \multicolumn{2}{c}{KonIQ10K}  \\
    & & SRCC & PLCC & SRCC & PLCC & SRCC & PLCC \\
    \midrule
    \multirow{2}{*}{HyperIQA} & Original & 0.8665 & 0.8883 & 0.8553 & 0.8716 & 0.8990 & 0.9190 \\
                              & QPT   & \textbf{0.8982}$_{+3.17\%}$ & \textbf{0.9061 }$_{+1.78\%}$ & \textbf{0.8743}$_{+1.90\%}$ & \textbf{0.8732}$_{+0.16\%}$ & \textbf{0.9066}$_{+0.76\%}$ & \textbf{0.9241}$_{+0.51\%}$ \\
    \midrule
    \multirow{2}{*}{TRes}     & Original & 0.7541 & 0.7918 & 0.8436 & 0.8793 & 0.8958 & 0.9097 \\
                              & QPT   & \textbf{0.7936}$_{+3.95\%}$ & \textbf{0.8037}$_{+1.19\%}$ & \textbf{0.8666}$_{+2.30\%}$ & \textbf{0.8882}$_{+0.89\%}$ & \textbf{0.9052}$_{+0.94\%}$ & \textbf{0.9150}$_{+0.53\%}$ \\
    \bottomrule
    \end{tabular}
    \label{tab:replace}
    \vspace{-0.3cm}
\end{table*}

\begin{table}[h]
	\centering
	\scriptsize
	\caption{Cross-dataset evaluation. Results of SRCC are reported.}
	\begin{tabular}{cc|ccc|c}
		\toprule  
		Train & Test & DBCNN & HyperIQA & CONTRIQUE & QPT \\ 
		\midrule  
		KIQ10k & CLIVE & 0.755 & 0.785 & 0.731 & \textbf{0.8208} \\
		KIQ10k & BID & 0.816 & 0.819 & - & \textbf{0.8247} \\
		CLIVE & BID & 0.762 & 0.756 & - & \textbf{0.8449} \\
		CLIVE & KIQ10k & 0.754 & \textbf{0.772} & 0.676 & 0.7494 \\
		\bottomrule  
	\end{tabular}
	\label{tab:cross}
\end{table}

\subsection{Comparison with SOTA BIQA Methods}

We report the performance of SOTA BIQA methods in \cref{tab:sota}. Generally, \textbf{our method pushes the current SOTA results a big step forward (using the same pre-trained weight generated by QPT)}, improving the best results to 0.8875 (+3.0\%) of SRCC and 0.9109 (+4.6\%) of PLCC in BID, 0.8947 (+4.0\%) of SRCC and 0.9141 (+2.4\%) of PLCC in CLIVE, 0.9271 (+1.1\%) of SRCC and 0.9413 (+1.3\%) of PLCC in KonIQ10K, 0.9250 (+0.8\%) of SRCC and 0.9279 (+0.7\%) of PLCC in SPAQ, and 0.5746 (+0.9\%) of SRCC and 0.6748 (+1.4\%) of PLCC in FLIVE. \textit{Notably, these results are achieved using an ordinary architecture of ResNet-50, showing the power of utilizing QPT models.} Besides, we also report the results of naive ResNet-50 (\ie, using ImageNet supervised pre-training).


Specifically, QPT models surpass traditional methods (\eg, NIQE \cite{DBLP:journals/spl/MittalSB13}, ILNIQE \cite{DBLP:journals/tip/ZhangZB15}, BRISQUE \cite{DBLP:journals/tip/MittalMB12} and BMPRI \cite{DBLP:journals/tbc/MinZGLY18}) that rely on hand-crafted features in large margins. Compared with earlier CNN-based methods (\eg, CNNIQA \cite{DBLP:conf/cvpr/KangYLD14}, WaDIQaM-NR \cite{DBLP:journals/tip/BosseMMWS18} and SFA \cite{DBLP:journals/tmm/LiJLJ19}), QPT shows the advantages of big data pre-training. For HyperIQA \cite{DBLP:conf/cvpr/SuYZZGSZ20} that utilized a self-adaptive hyper network architecture to estimate content understanding, perception rule learning and quality predicting, QPT-ResNet50 obtains higher results without fine-designed architectures. Compared with current SOTA Transformer-based methods (\eg, MUSIQ \cite{DBLP:conf/iccv/KeWWMY21} and TReS \cite{DBLP:conf/wacv/GolestanehDK22}), QPT models still outperform them, showing the effectiveness of quality-ware pretext training.

\textbf{Due to the flexibility of QPT, it can be easily integrated with current SOTA methods by replacing their pre-trained weights.} We reproduced two SOTA methods (\ie, HyperIQA and TReS) based on their official open-source code, giving the results in \cref{tab:replace}. The results of the comparison are based on a consistent dataset partitioning and training hyper-parameters, following official implementation. It can be seen that QPT further boosts SOTA methods effectively (\eg, a even better result with 0.8982 of SRCC in BID can be achieved), showing satisfactory generalization ability. We further give the corss-dataset evaluation in \cref{tab:cross}. Compared with the most competitive methods, QPT takes the lead in three scenarios.

\subsection{Ablation Studies}

\begin{table}[t]
    \centering
    \footnotesize
    \caption{Impact of data amount for the pretext task of QPT, using different percentages of the ImageNet dataset, in BID and CLIVE.}
    \begin{tabular}{c|cc|cc}
    \toprule
    \multirow{2}{*}{Percentage of ImageNet} & \multicolumn{2}{c|}{BID} & \multicolumn{2}{c}{CLIVE}  \\
    & SRCC & PLCC & SRCC & PLCC \\
    \midrule
    20\% & 0.8311 & 0.8305 & 0.8115 & 0.8203 \\
    50\% & 0.8760 & 0.8973 & 0.8868 & 0.9007 \\
    100\% & \textbf{0.8875} & \textbf{0.9109} & \textbf{0.8947} & \textbf{0.9141} \\
    \bottomrule
    \end{tabular}
    \label{tab:data_scale}
    \vspace{-0.3cm}
\end{table}

\noindent
\textbf{Impact of data amount for QPT.} To evaluate the effectiveness of the quality-aware pretext task, we use 20\%, 50\% and 100\% percentages of the ImageNet dataset, analyzing the impact on the number of used data for downstream results. As given in \cref{tab:data_scale}, the larger the number used in the pre-training process, the better the performance on the downstream IQA datasets. 
Still, it is quite a surprise that our method obtains SOTA results with only 50\% of the data.
Furthermore, training larger-scale data is an open question and can be investigated in future work, where QPT may benefit from JFT-300M \cite{DBLP:conf/iccv/SunSSG17}.

\begin{table}[t]
    \centering
    \footnotesize
    \caption{Impact of encoder capacity for QPT, using different series of networks, in BID and CLIVE.}
    \begin{tabular}{c|c|cc|cc}
    \toprule
    \multirow{2}{*}{Network} & \multirow{2}{*}{Params} & \multicolumn{2}{c|}{BID} & \multicolumn{2}{c}{CLIVE}  \\
    & & SRCC & PLCC & SRCC & PLCC \\
    \midrule
    MBViTV2-100 & 4.90 & 0.8746 & 0.8799 & 0.8640 & 0.8914 \\
    MBViTV2-150 & 10.59 & 0.8782 & 0.8976 & 0.8702 & 0.8970\\
    MBViTV2-200 & 18.45 & \textbf{0.8821} & \textbf{0.8960} & \textbf{0.8894} & \textbf{0.9006} \\
    \midrule
    ResNet-18 & 11.69 & 0.8679 & 0.8805 & 0.8610 & 0.8795 \\
    ResNet-34 & 21.80 & 0.8742 & 0.9001 & 0.8780 & 0.8871 \\
    ResNet-50 & 25.56 & \textbf{0.8875} & \textbf{0.9109} & \textbf{0.8947} & \textbf{0.9141} \\
    \bottomrule
    \end{tabular}
    \label{tab:model_scale}
    \vspace{-0.3cm}
\end{table}

\noindent
\textbf{Impact of encoder capacity for QPT.} To verify the effect of model capacity on feature modeling, we select two series of networks for analysis, including the widely-used CNN architectures of ResNet and an emerging Transformer-based architectures of MobileViTV2 \cite{DBLP:journals/corr/abs-2206-02680}. As shown in \cref{tab:model_scale}, increasing the model size can help improve the ability of feature representation, resulting in better downstream performance. 
It should be pointed out that large models will also increase the time consumption of the pre-training phase (\ie, ResNets of \{53h, 61h, 75h\}, MBViTV2s of \{66h, 92h, 109h\}, respectively). We did not use larger networks out of tradeoffs, which can be tried in future work.

\begin{table}[t]
    \centering
    \scriptsize
    \caption{Impact on the composition of negative samples.}
    \vspace{-0.2cm}
    \begin{tabular}{cc|cc|cc}
    \toprule
    \multirow{2}{*}{Inter} & \multirow{2}{*}{Intra} & \multicolumn{2}{c|}{BID} & \multicolumn{2}{c}{CLIVE}  \\
        & & SRCC & PLCC & SRCC & PLCC \\
        \midrule
        \checkmark & \checkmark & \textbf{0.8875} & \textbf{0.9109} & \textbf{0.8947} & \textbf{0.9141} \\
        \midrule
        \checkmark & $\times$   & 0.8702 & 0.9051 & 0.8721 & 0.8994 \\
        $\times$ & \checkmark   & 0.8399 & 0.8357 & 0.8084 & 0.8142 \\
        $\times$ & $\times$     & 0.7078 & 0.7731 & 0.7495 & 0.7493 \\
        \bottomrule
    \end{tabular}
    \label{tab:negative}
    \vspace{-0.1cm}
\end{table}

\noindent
\textbf{Impact of different compositions of negative samples.} To evaluate the effectiveness of negative samples and the proposed QC Loss, we give the results in \cref{tab:negative}. The highest result is achieved when both types of negative samples are present. And the absence of inter-sample negative pairs leads to a larger performance drop. 
This is due to the fact that compared with intra-sample negative pairs (\ie, $K-1$), the number of inter-sample ones (\ie, $N\cdot K$), which contain both different content and degradation types, is larger.

\begin{figure}[t]
  \centering
  \begin{subfigure}{0.47\linewidth}
    \includegraphics[width=\columnwidth]{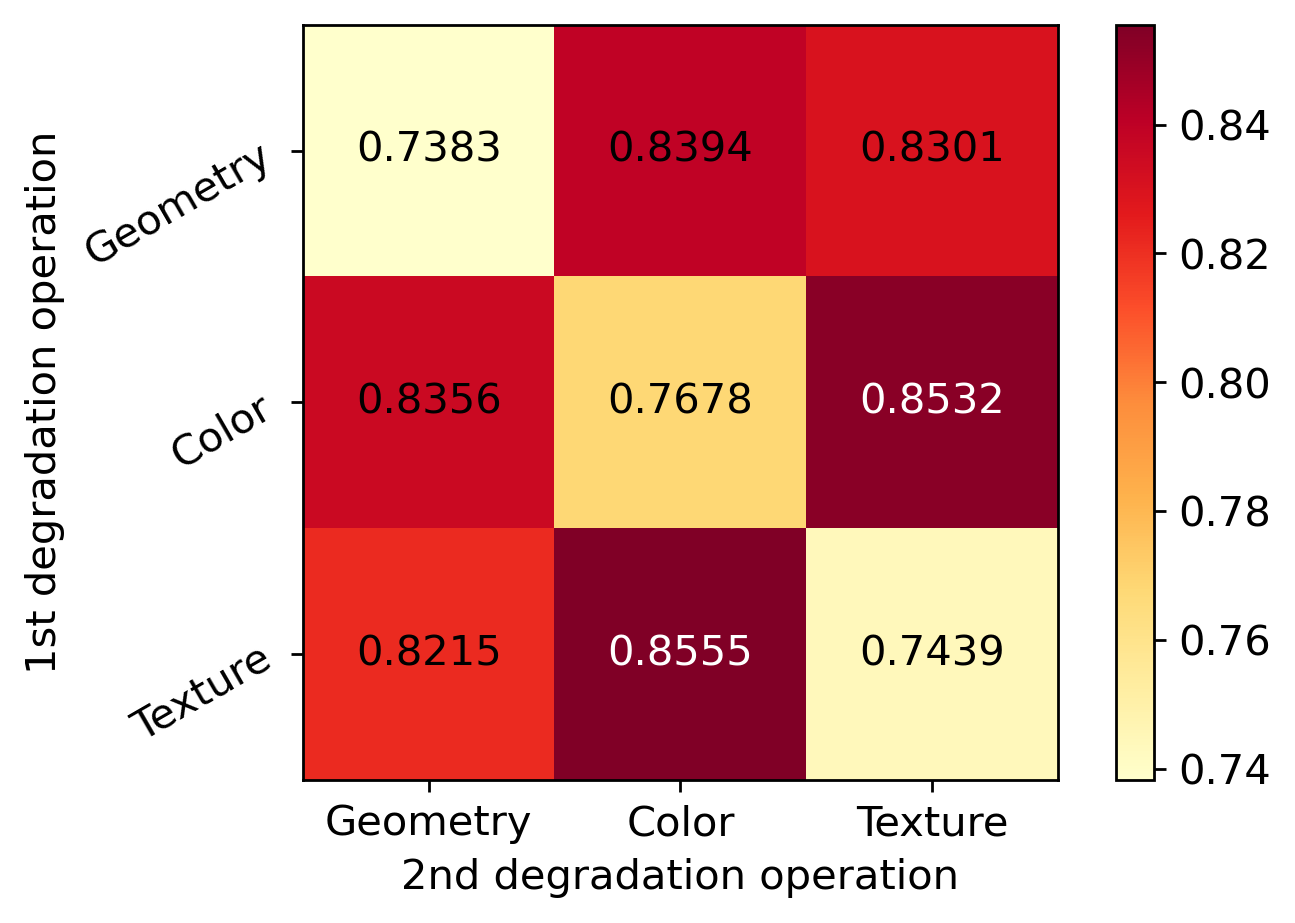}
    \caption{Results in BID}
    \label{fig:bid}
  \end{subfigure}
  \begin{subfigure}{0.47\linewidth}
    \includegraphics[width=0.98\columnwidth]{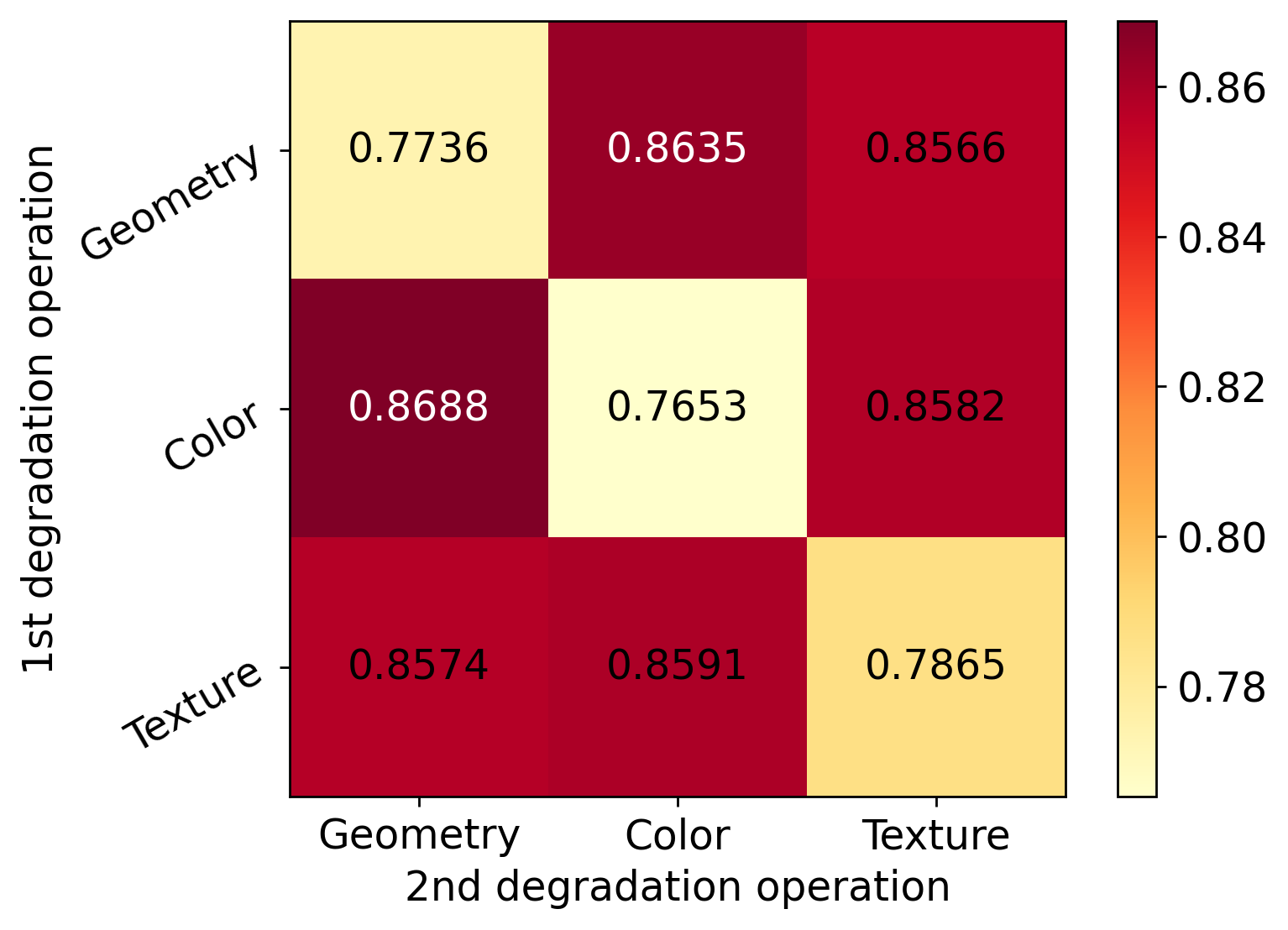}
    \caption{Results in CLIVE}
    \label{fig:clive}
  \end{subfigure}
  \caption{Evaluation results of SRCC under different fixed compositions of degradations in a first-order manner. The diagonal entries denote single class degradation type, and off-diagonals represent composition of two-class degradation applied sequentially.}
  \label{fig:heatmap}
  \vspace{-0.3cm}
\end{figure}

\noindent
\textbf{Ablation on the degradation space.} Experiments are performed to evaluate the impact of different degradations and compositions. In \cref{fig:heatmap}, we give the results under different compositions of degradation types in a first-order manner. Trying all combinations is impractical due to the huge degradation space. Experiments are performed in a coarse-grained manner. The results show that there is a clear effect difference between the degradation types of a fixed form. More flexible and larger space are needed to cover different downstream scenarios. We further evaluate the strategy of composition as shown in \cref{fig:aug}. In \cref{tab:space}, a combination of the three strategies yielded the best results.

\begin{table}[t]
    \centering
    \scriptsize
    \caption{Ablation on the degradation space (different formulation type in \cref{equ:degradation} of complicated degradation models) in BID.}
    \vspace{-0.1cm}
    \begin{tabular}{ccc|cc}
        \toprule
        skip & shuffle & two-order & SRCC & PLCC \\
        \midrule
        \checkmark & \checkmark & \checkmark & \textbf{0.8875} & \textbf{0.9109} \\
        \checkmark & \checkmark & $\times$   & 0.8828 & 0.9075 \\
        \checkmark & $\times$ & $\times$     & 0.8745 & 0.9029 \\
        $\times$ & $\times$ & $\times$       & 0.8626 & 0.8864 \\
        \bottomrule
    \end{tabular}
    \label{tab:space}
\end{table}

\subsection{Comparisons with other pretext tasks}

We further compare QPT with other pretext methods, including train-from-scratch (\ie, \textit{w/o} pre-training weights), supervised and MoCo. We extract features from the encoder output for both fine-tuning and linear probing evaluations.

\begin{table}[t]
    \centering
    \scriptsize
    \caption{Comparisons of \textbf{linear probing} evaluation using different pretext tasks in BID and CLIVE.}
    \vspace{-0.1cm}
    \begin{tabular}{c|cc|cc}
    \toprule
    \multirow{2}{*}{Pretext task} & \multicolumn{2}{c|}{BID} & \multicolumn{2}{c}{CLIVE}  \\
    & SRCC & PLCC & SRCC & PLCC \\
    \midrule
    Supervised & 0.7678 & 0.7860 & 0.7128 & 0.7668 \\
    MoCo & 0.7367 & 0.7357 & 0.6796 & 0.7154 \\
    QPT & \textbf{0.8242} & \textbf{0.8175} & \textbf{0.7613} & \textbf{0.7938} \\
    \bottomrule
    \end{tabular}
    \label{tab:linear}
\end{table}

\noindent
\textbf{Linear probing evaluation.} In this way, the weights of encoder are frozen for feature extraction directly. As given in \cref{tab:linear}, QPT obtains the best performance, surpassing MoCo by 8.75\% of SRCC in BID and 8.17\% in CLIVE.

\begin{table}[t]
    \centering
    \scriptsize
    \caption{Comparisons of \textbf{end-to-end fine-tuning} evaluation using different pretext tasks in BID and CLIVE.}
    \vspace{-0.1cm}
    \begin{tabular}{c|cc|cc}
    \toprule
    \multirow{2}{*}{Pretext task} & \multicolumn{2}{c|}{BID} & \multicolumn{2}{c}{CLIVE}  \\
    & SRCC & PLCC & SRCC & PLCC \\
    \midrule
    \textit{w/o} & 0.7078 & 0.7731 & 0.7495 & 0.7493 \\
    Supervised & 0.8423 & 0.8521 & 0.8527 & 0.8807 \\
    MoCo & 0.8531 & 0.8724 & 0.8494 & 0.8615 \\
    QPT & \textbf{0.8875} & \textbf{0.9109} & \textbf{0.8947} & \textbf{0.9141} \\
    \bottomrule
    \end{tabular}
    \label{tab:finetune}
    \vspace{-0.2cm}
\end{table}

\noindent
\textbf{End-to-end fine-tuning evaluation.} Linear probing has been a popular protocol in the past few years. However, it misses the opportunity of pursuing strong but nonlinear features which is indeed a strength of deep learning \cite{DBLP:conf/cvpr/HeCXLDG22}. We also give the fine-tuning results in \cref{tab:finetune}. QPT still achieves the best results. Generally, higher results can be obtained than linear probing. These results further prove the effectiveness of QPT over instance-level discrimination or supervised classification pretext tasks in IQA scenarios.

\section{Conclusion}

In this paper, we proposed the QPT to generate pre-trained models for downstream BIQA tasks, alleviating the obstacle of insufficient annotated data. By introducing diverse degradation types and compositions, we construct a degradation space that contains $2\times 10^7$ possible degradations, to simulate more complicated and realistic distorted images. To constrain the learning process, we proposed the QC-Loss that treats patch pairs extracted from the same degraded images as positive ones. And other pairs are noted as negative ones, including degradation-based and content-based ones. After pre-trained on ImageNet using QPT, BIQA models obtain significant improvements on five BIQA benchmark datasets. Moreover, QPT can be easily integrated with current SOTA methods by replacing their pre-trained weights, showing good generalization ability.

\clearpage
{\small
\bibliographystyle{ieee_fullname}
\bibliography{egbib}
}

\end{document}